\documentclass[10pt, conference, compsocconf]{IEEEtran}
\pdfoutput=1

\ifCLASSINFOpdf
\usepackage[pdftex]{graphicx}
\else
\fi
\usepackage[caption=false,font=footnotesize]{subfig}
\begin{document}
%
\title{Automatic Image Registration in Infrared-Visible Videos using Polygon Vertices}




%

\author{\IEEEauthorblockN{ Tanushri Chakravorty,  Guillaume-Alexandre Bilodeau}
\IEEEauthorblockA{LITIV Lab\\
\'{E}cole Polytechnique de Montr\'{e}al\\
Montr\'{e}al ,QC, Canada\\
Email: tanushri.chakravorty@polymtl.ca, gabilodeau@polymtl.ca}
\and
\IEEEauthorblockN{Eric Granger}
\IEEEauthorblockA{LIVIA\\
\'{E}cole de technologie sup\'{e}rieure\\
Montr\'{e}al , QC, Canada\\
Email: Eric.Granger@etsmtl.ca}
}


\maketitle

\begin{abstract}
In this paper, an automatic method is proposed to perform image registration in visible and infrared pair of video sequences for multiple targets. In multimodal image analysis like image fusion systems, color and IR sensors are placed close to each other and capture a same scene simultaneously, but the videos are not properly aligned by default because of different fields of view, image capturing information, working principle and other camera specifications. Because the scenes are usually not planar, alignment needs to be performed continuously by extracting relevant common information.
In this paper, we approximate the shape of the targets by polygons and use affine transformation for aligning the two video sequences. After background subtraction, keypoints on the contour of the foreground blobs are detected using DCE (Discrete Curve Evolution)technique. These keypoints are then described by the local shape at each point of the obtained polygon. The keypoints are matched based on the convexity of polygon's vertices and Euclidean distance between them. Only good matches for each local shape polygon in a frame, are kept. To achieve a global affine transformation that maximises the overlapping of infrared and visible foreground pixels, the matched keypoints of each local shape polygon are stored temporally in a buffer for a few number of frames. The matrix is evaluated at each frame using the temporal buffer and the best matrix is selected, based on an overlapping ratio criterion. Our experimental results demonstrate that this method can provide highly accurate registered images and that we outperform a previous related method.

\end{abstract}


%
\IEEEpeerreviewmaketitle

\section{Introduction}
\begin{figure}[!ht]%
\centering
\subfloat[]
{\includegraphics[width=2.5in]{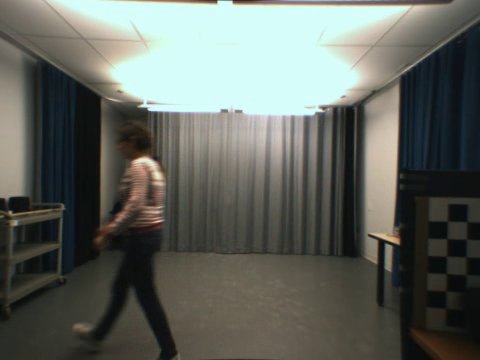}
\label{fig:fig1-a}}\\
\subfloat[]
{\includegraphics[width=2.5in]{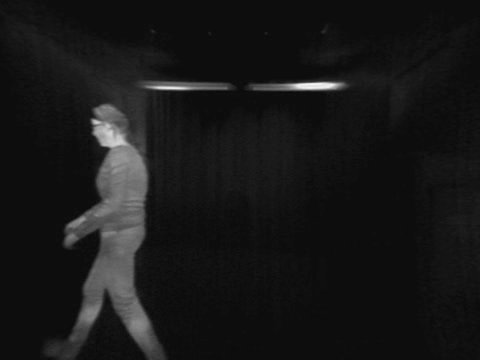}
\label{fig:fig1-b}}
\caption{Different image information captured of a same scene by (a) visible camera (b) and infrared camera, respectively. Note the absence of the striped texture of the shirt in infrared.}%
\label{fig:fig1}%
\end{figure}

Nowadays there has been an increasing interest in infrared-visible stereo pairs in video surveillance because both sensors complement each other. This has led to the development of variety of applications ranging from medical imaging, computer vision, remote sensing, astrophotography etc. to extract more information about an object of interest in an image. Visible camera provides information about the visual context of the objects in the scene, but under poor light conditions only limited information is captured. On the other hand infrared provides enhanced contrast and rich information about the object when there is less light, especially in the dark environment. Example of such different capturing information is shown in Figure \ref{fig:fig1}. Therefore, to benefit from both the modalities, it is required to extract information from both the capturing sources for which image registration is a necessary step.

Infrared-visible image registration is a very challenging problem since the thermal and visible sensors capture different information about a scene \cite{trajec}. The infrared captures the heat signature emitted by objects, while the visible captured the light reflected by objects. Due to this difference, the correspondence between the visible and the infrared is hard to establish as local intensities or textures do not match, as can be seen in the Figure \ref{fig:fig1}.

Therefore, to detect and match common features such as appearance, shape etc. in the image captured by both cameras become very difficult. This problem further becomes more challenging with the increase in number of targets in the scene, as the complexity of the system increases. Hence, in order to perform more accurate infrared-visible image registration for multiple targets such as people, we utilise a method based on keypoint features on target boundaries and temporal information between matched keypoint pairs to calculate the best affine transformation matrix.

Hence in this method to establish feature point correspondence, boundary regions described by visible and infrared videos are considered as noisy polygons and the aim is to calculate the correspondence between the vertices of these polygons. This information can be further used for fusion of infrared and visible image to improve object detection and tracking, recognition etc. Since infrared is relatively invariant to changes in illumination, it has the capability for identifying object under all lighting conditions, even in total darkness. 

Hence, the worthy information provided by infrared images is a potential for the development of surveillance applications. In this paper, both the cameras are placed parallel and nearby in a stereo pair configuration, i.e. the cameras observe a common viewpoint \cite{trajec}. Note, that we do not assume that the scene is planar, but we do assume that all targets are about in the same plane. That is, the group of targets are moving together through different planes throughout the videos. Our method can be generalized to many targets in different planes.\\

This paper extends the work of \cite{socheat}. The contributions of this paper are:
\\
\begin{enumerate}
\item We improve registration accuracy over the state of the art, since we consider each polygon obtained in a frame separately, which results in more precise keypoint matches and less outliers. If we consider all the polygons at a single time for finding matches as in \cite{socheat}, much ambiguity arises and the method is only limited to targets in a single plane at a time.
\\
\item By considering each polygon separately, our method can be generalized to a scene with targets appearing simultaneously in many depth planes.
\\
\item We propose a more robust criterion to evaluate the quality of a transformation and use the overlapping ratio of the intersection of foreground pixels of the infrared and visible images over the union of these foreground pixels. This allows us to update the current scene transformation matrix only if the new one improves on the accuracy.  
\end{enumerate}

This paper is structured as follows. Section \ref{sec:previous} discusses the recent work done in the field. The proposed algorithm is presented in Section \ref{sec:methodology}. Section \ref{sec:evaluation} presents the registration results and evaluation accuracy of the algorithm with the ground-truth. Finally Section \ref{sec:conc} summarises the paper.

\section{Previous Work}
\label{sec:previous}

Image registration technique has been mainly applied in thermography and multimodal analysis, in which a particular scene is captured using visible and infrared sensor from different viewpoints to extract more information about it. For extracting common information, often image regions are considered in both the infrared and the visible by using a similarity measure like LSS (Local Self Similarity) \cite{lss} or MI (Mutual Information) \cite{survey}. LSS and MI are easy to compute over regions, but the procedure becomes slow when used as a feature for image registration, particularly when there are many targets and registration has to be computed at every frame. Therefore, features like boundaries \cite{edge},\cite{edge2}, edges or connected edges, are one of the most popular approaches for extracting common information in this scenario.

Features such as corners, are also used to perform matching in image registration \cite{corners}. The corners are detected on both the visible and infra red images and then similarity is measured using Hausdorff distance \cite{haus}. Furthermore, the features such as line segments and virtual line intersections have also been used for registration \cite{line}. To find correspondence between frames, recent methods like blob tracking \cite{blob} and trajectory computation \cite{trajec} have also been used. But these methods are complex, since they need many trajectories to achieve a good transformation matrix and hence more computation. Also, they only apply to planar scenes.

Recently, Sonn et al. \cite{socheat} have presented a method based on polygonal approximation DCE (Discrete Curve Evolution) \cite{dce} of foreground blobs for fast registration. The method gives promising results, but it is limited in precision because it only considers the scene globally. We extend this method by considering each target individually for better matching precision. This will in turn allow calculating a transformation for each individual target. We also improved transformation matrix selection. Therefore, in our proposed method, we extract features such as keypoints on each contour. The advantage of using keypoint features is that they are simple and easy to calculate. Also, to have more matches, the keypoints are stored in a temporal buffer that is continually renewed after a fixed number of frames, thus resulting in a better transformation for the video sequence. Our experiments show that we obtain better results as compared to a recent method used for registering infrared-visible video sequences.

\section{Methodology} 
\label{sec:methodology}

\begin{figure}[!ht]
\centering
\includegraphics[width=3in]{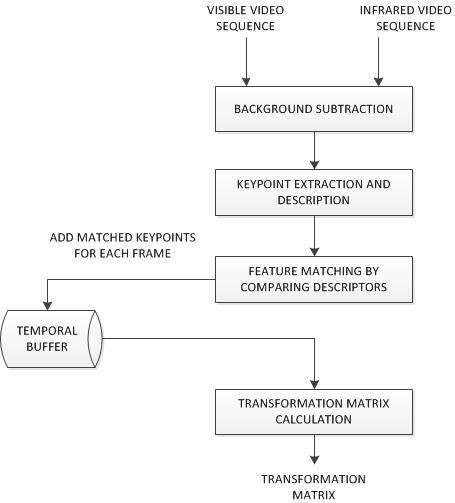}
\caption{System Block Diagram}
\label{fig:sysdia}
\end{figure}

Our proposed method consists of four main steps as shown in Figure \ref{fig:sysdia}. The first step of the algorithm performs background subtraction according to the method explained in \cite{Shoushtarian20055} to get the foreground blob regions (contours). The second step of the algorithm performs feature detection and extraction using DCE (Discrete Curve Evolution) technique \cite{dce} which outputs significant keypoints on the contours. For feature description, the significant keypoints detected are described by the local polygon shapes computed at each keypoint. The third step of the algorithm performs feature matching by comparing the feature descriptors obtained at the previous step using similarity measures as described later in the Section \ref{sec:fm} of the paper. The corresponding matches are saved in a temporal buffer. This temporal buffer allows accumulating matches from recent observations to reduce noise and improve registration precision. The fourth and final step of the algorithm calculates the homography matrix based on the result of the matched descriptors stored in the temporal buffer in the previous step and hence we obtain a transformation matrix at the end of our algorithm. This process is applied at every frame since the target is assumed to move throughout different planes. The buffer is refreshed after a few frames so as to keep the latest matched keypoints, which helps in estimating the recent homography which best transforms the scene. All the keypoints in the temporal buffer are used to calculate the transformation matrix.

\subsection{Background Subtraction}
The objective of the background subtraction method is to identify the foreground objects in an image. Techniques like frame differencing \cite{bas} computes the difference between two consecutive frames depending on a threshold. But this technique may not be useful for images having fast motion since the method relies on a global threshold. Also, it would not give complete object contours. This is why a proper background subtraction technique needs to be used. Better background subtraction methods will give better registration accuracy. In this work we have used a simple background subtraction technique based on temporal averaging \cite{Shoushtarian20055}. The advantage of this method is that it is fast and the algorithm selectively segments foreground blobs from the background frames. This method produces correct background image for each input frame and is able to remove ghosts and stationary objects in the background image efficiently. Any other method could be used.

\subsection{Feature Extraction}
After the background subtraction step, which gives the foreground blobs, an adapted DCE (Discrete Curve Evolution) \cite{IR} is used to detect salient points on the infrared-visible foreground blobs. These points represent points of visual significance along the contour and the branches connected to these points represent the shape significance. Branches that end at a concave point are discarded and the output of the algorithm converges to a convex contour. The method also filters out noisy points along the contours. Hence, the method approximates each contour of a foreground blob to a polygon. The significant number of vertices or points on each contour that are determined by the DCE algorithm can be set by the user, which is set as 16 in our case. Before the matching process, the internal contours or holes are also removed from each contour. 

\subsection{Feature Description}
Each significant keypoint obtained by DCE is described by the local shape of the polygon at that point \cite{socheat}. The properties of the local polygon vertices like convexity and angle of the polygon are used to describe the keypoints. Hence a feature descriptor is composed of two feature vectors with two components viz., convexity and angle of the polygon. For example, consider three consecutive points ${P1}$, ${P2}$, ${P3}$ on a contour in clockwise direction. The convexity of a single point is defined by the normal vector to the remaining other points and can be defined by equation:
\begin{equation}
\vec{n} = \vec{P_{12}} \times \vec{P_{23 }} \ ,
\label{eq:findConvexity}
\end{equation}

where $\vec{n}$ is the normal vector, $\vec{P_{12}}$ is a vector from $P_1$ to $P_2$ and $\vec{P_{23}}$ is a vector from $P_2$ to $P_3$. Here each keypoint is supposed to have three dimensional coordinates given by ${(x,y,0)}$. After the cross product, $\vec{n}$ will contain the value of ${z}$ coordinate. This ${z}$ value is evaluated for determining convexity of keypoints. If its value is less than zero, the polygon is considered convex, else it is concave. Only those contours are kept for further processing which suffice this criterion of convexity.

For computation of angle $\theta$ for a keypoint, it is calculated by the following equation:
\begin{equation}
\theta = \cos^{-1}{ \left( \frac{\left|{\vec{P_{21}}}\right|^2 + \left|{\vec{P_{23}}}\right|^2 - \left|{\vec{P_{13}}}\right|^2}{2 * \left|\vec{P_{21}}\right| * \left|\vec{P_{23}}\right|} \right)} \ 
\label{eq:findAngle}
\end{equation}

where $\theta$ is the angle formed between $\vec{P_{21}}$ and$\vec{P_{23}}$.

\subsection{Feature Matching}
\label{sec:fm}

To find the correspondence between the keypoints, each polygon is analysed separately one by one, in both visible and infrared foreground regions. This results in a larger number of pair of matches as compared to the method in \cite{socheat}, where all the polygons are analysed at a single time for the whole image. This also allows us to register each target individually if required. Therefore, in this work, we have to find both the best polygon matches and the best keypoint matches. The features are matched by comparing feature descriptors obtained in the previous step. The similarity conditions like convexity, euclidean distance and difference of angles between the feature descriptors determine the percentage of matching between the polygons. Therefore, the matching criteria is given by the following conditions \cite{socheat}:\\

\begin{enumerate}

\item Convexity, $c$: It is calculated using Eq. \ref{eq:findConvexity}. Only those points are kept which satisfy the criteria of having the value of ${z}$ greater than zero.
\\
\item $E_d$: It is the euclidean distance between two keypoints.
\\
\begin{equation}
E_d = \left|P_I - P_V\right| \ 
\label{eq:findDistance}
\end{equation}
\item $E_\theta$:  The difference between the two keypoint angles.
\begin{equation}
E_\theta = \left|\theta_I - \theta_V \right|\ 
\label{eq:findAngleError}
\end{equation}

\end{enumerate}

The pair of keypoints from visible and infrared images which fulfil the convexity criteria, $c$, Eq. \ref{eq:findConvexity} are kept. Then the euclidean distance, $E_d$  between the two keypoints, and the difference between the two keypoint angles, $E_\theta$ is calculated using Eq. \ref{eq:findDistance} and Eq. \ref{eq:findAngleError}, respectively. The threshold for Euclidean distance is set as ${E_d}_{Max}$, and for maximum angle error, as ${E_\theta}_{Max}$.

Only those pair of keypoint is kept for which $E_d \leq {E_d}_{Max}$ and $E_\theta \leq {E_\theta}_{Max}$ is true and the other pairs of keypoints are rejected. If there are keypoints in infrared for which there is more than single match in the visible, the best match for those keypoints is selected by a Score criteria as mentioned in \cite{socheat}.  The Score, ${S}$ is calculated as:\\  

\begin{equation}
S = \frac{\alpha E_d}{{E_d}_{Max}} + \frac{E_\theta}{{E_\theta}_{Max}} 
\label{eq:findScore}
\end{equation}

Additionally, contrarily to \cite{socheat}, we only keep matches that are on the best matching pairs of polygons. The matched keypoints for each polygon in both visible and infrared image are saved in a temporal buffer of matches, since it might not be possible to have a significant number of matched keypoints, when a single frame is considered. Therefore, the temporal buffer stores the matched keypoints for a few numbers of frames and is renewed with new keypoints. The temporal buffer gets filled in a similar to first-in-first-out-fashion. This technique helps to attain a significant number of matched keypoints, which will result in a more meaningful and accurate calculation of homography matrix. One or more temporal buffer of matches can be used. To register each object individually, a temporal buffer should be attributed to each object. Tracking of objects may be required to distinguish the different temporal buffers. In this work, to test our proposed improvements we have used a single temporal buffer, and assume that all objects move together in a single plane. We will see later on, that even in this case, considering matches on a polygon by polygon basis, improves accuracy because matching ambiguities are greatly reduced.

\subsection{Transformation matrix calculation and matrix selection}
The pairs of matched keypoints stored in the temporal buffer are used to determine an accurate transformation matrix for every frame of the whole video scene. The temporal buffer should not have a too long temporal extent as the target will gradually move from on plane to another. Therefore, the saved matches work in a FIFO (first-in-first-out) method. With this technique, only the last few frames are required to calculate the transformation matrix. The matrix thus obtained is more accurate, since the last saved pair of matches in the temporal buffer resemble more to the polygons that were present in the last number of frames in the video sequence, which are about in the same plane.

To calculate the homography and filter the outlier points in the buffer, the RANSAC (RANdom SAmple Consensus) is used \cite{Fischler:1981:RSC:358669.358692}. The matrix is calculated for each frame and the best matrix is saved and applied to the infrared foreground frame, which becomes a transformed infrared foreground frame. For selecting the best matrix, the blob overlap ratio is evaluated at every frame. It is given by:
\begin{equation}
BR=\frac{A_I \cap A_V}{A_I \cup A_V} ,
\label{eq:findRatio}
\end{equation}
where, ${A_I}$ and ${A_V}$ are the foreground regions of the transformed infrared and visible blobs respectively. Only that matrix is selected for which the overlap ratio, ${BR}$, is close to 1. This improved selection criterion contributes to an improvement in the precision.

\section{Evaluation of image registration accuracy}
\label{sec:evaluation}
We have used \textit{Alignment Error} to measure the accuracy of our method. In this, the mean square error is evaluated at the ground-truth points selected in visible and infrared video frames. The alignment error gives a measure that how different is the transformation model obtained by image registration method from the ground-truth. 

To test our new method, we calculated a single transformation matrix at each frame as we are using a single temporal buffer in the current implementation of our method. This choice has the benefit that allows us to compare our proposed method with the one of Sonn et al. \cite{socheat}, which only considers the scene globally at each frame. To allow comparison with the state-of-the-art, we applied the method of Sonn et al. \cite{socheat} on our video sequences using their default parameters (which was provided to us by the authors). Our method was tested on selected frames distributed throughout the video. The mean error was calculated for points selected over regions of persons. The number of persons present in the scene varies from 1 to 5. For the selected test frames, they are approximately together in a single plane (see Figure \ref{fig:etapes-d}). We have detailed the results for the various number of people, and thus, the various numbers of planes that are potentially at each frame.

For each sequence, two tests were done. One is for the 30 frame buffer size and the other is for 100 frame buffer size respectively. Both methods were tested with the same buffer size. The parameters used for our method are ${E_d}_{Max}=65$ and ${E_\theta}_{Max}=40$ degrees respectively.

\begin{table}
\begin{center}
\caption{Mean Registration error for 1-5 persons in the visible and infra-red video sequence for 30 and 100 temporal buffer size respectively. Mean Registration error: $E = \sqrt{{E_x}^2 + {E_y}^2}$ compared to the ground-truth.} 
\begin{tabular}{|l|l|l|l| p{2cm}|}
    \hline
    ~               & \# Person & Our Method & Sonn et al. \cite{socheat} \\ \hline
    30 Buffer Size  & 1        & \textbf{0.8952}     & 2.038      \\ \hline
    ~               & 2        & \textbf{1.6361}     & 3.2238     \\ \hline
    ~               & 3        & \textbf{1.0823}     & 5.6769     \\ \hline
    ~               & 4        & \textbf{0.9385}     & 5.5574     \\ \hline
    ~               & 5        & \textbf{1.2071}     & 4.4275     \\ \hline
    100 Buffer Size & 1        & \textbf{0.6084}     & 2.3522     \\ \hline
    ~               & 2        & \textbf{1.3724}     & 2.8458     \\ \hline
    ~               & 3        & \textbf{2.0170}     & 4.1089     \\ \hline
    ~               & 4        & \textbf{1.0538}     & 4.6247     \\ \hline
    ~               & 5        & \textbf{1.1354}     & 4.5206     \\ \hline
    \end{tabular} 
\end{center}
\label{table:table1}
\end{table}

The table \ref{table:table1} shows the mean registration error for 1-5 persons at the selected frames in the video and shows that our method outperforms the previous work \cite{socheat} for all the various number of people, in the video sequence for both buffer size. Since we have considered each contour separately one by one, in the video sequences, we have more number of keypoints or features, which helps in better matching. The remaining noisy keypoints are filtered by the RANSAC algorithm.

\begin{figure}[!ht]%
\centering
\subfloat[]
{\includegraphics[width=1.5in,height=1.41in]{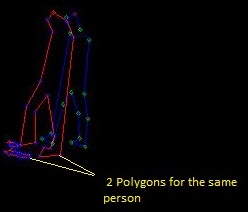}
\label{fig:fig3-a}}%
\subfloat[]
{\includegraphics[width=1.5in]{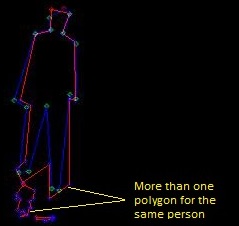}
\label{fig:fig3-b}}
\caption{(a) and (b) More than one polygon for a single person, after Background Subtraction step. Our method filters out the contradictory matches between polygons, since we consider each polygon individually.}%
\label{fig:fig3}%
\end{figure}

\begin{figure*}
\centering
\subfloat[]
{\includegraphics[width=2.5in]{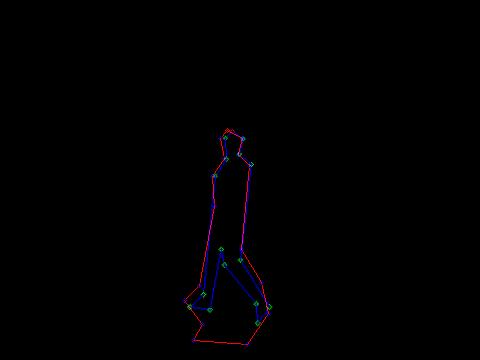}
\label{fig:etapes-a}} 
\subfloat[]
{\includegraphics[width=2.5in]{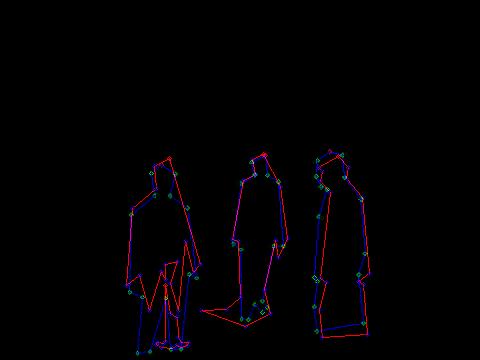}
\label{fig:etapes-b}}\\
\subfloat[]
{\includegraphics[width=2.5in]{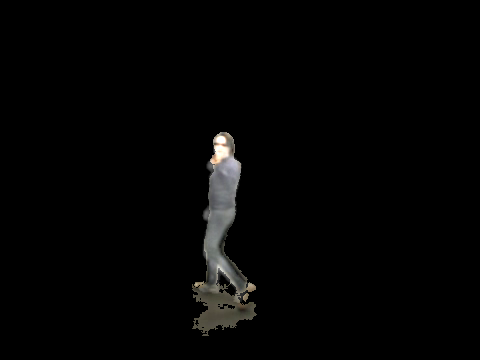}
\label{fig:etapes-c}} 
\subfloat[]
{\includegraphics[width=2.5in]{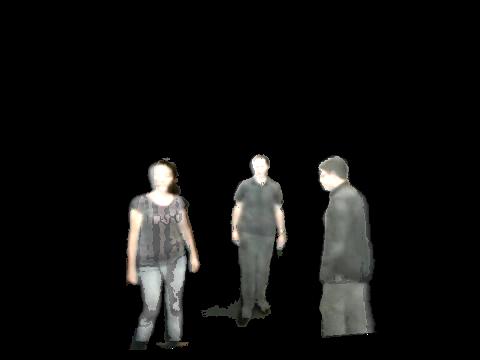}
\label{fig:etapes-d}}\\
\subfloat[]
{\includegraphics[width=2.5in]{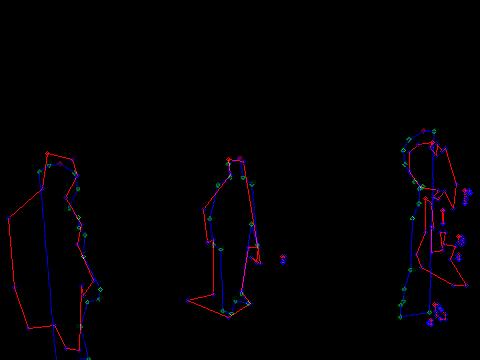}
\label{fig:etapes-e}}
\subfloat[]
{\includegraphics[width=2.5in]{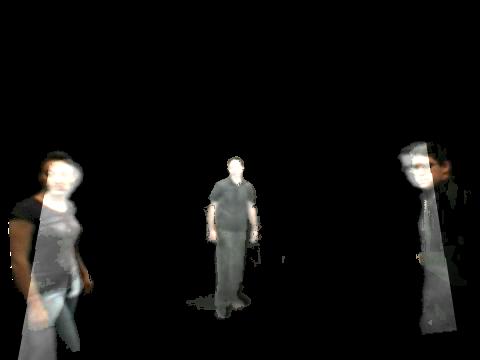}
\label{fig:etapes-f}}
\caption[example] 
{ \label{fig:etapes} 
Transformation results. Figures 4.(a) and (b) show the overlapping regions of the transformed IR polygons over the Visible polygons and Figures 4.(c) and (d) show the overlapping regions of the transformed IR over the foreground visible regions. Figures 4.(e) and (f) show the results when the persons are not in the same image plane.}
\end{figure*}

The fact that our new method outperforms the method of Sonn at al.\cite{socheat} for even only one person is significant. This can be explained by the background subtraction that is not perfect and results in many polygons even in the case of one person as shown in Figure \ref{fig:fig3}. Considering each polygon individually allows us to select better matches and remove contradictory information between polygon matches.

This shows that matching polygon vertices globally is error prone, as the local shape of the polygon vertices are not necessarily unique. Thus, we should ensure that the matching vertices are matched as a group from the same pair of polygons. Furthermore, because our transformation matrix criterion is better, we also update more adequately the current transformation matrix. The result shows that the error varies between 0.5 and 2 pixels. Since the buffer size has a small impact on matching between keypoints, we can choose the buffer size depending on the application. Figure \ref{fig:etapes} shows transformation results. It can be noted that the chances of registration error increases in cases where the people are not exactly in the same depth plane (see Figure \ref{fig:etapes-e} and \ref{fig:etapes-f}). For such cases. we the can improve the results by calculating more than one transformation matrix for each person.

\section{Summary and conclusions}
\label{sec:conc}
We have presented an alternative approach to other image registration methods, such region-based, frame-by-frame keypoints-based and trajectory-based registration methods that works for visible and infrared stereo pairs. The method uses a feature based on polygonal approximation and a temporal buffer filled with matched keypoints. The results show that our method outperforms \cite{socheat}, for every tested sequence. As we have more considered each contour locally one by one in the video sequence, we obtain more features and hence more matches. To obtain the best transformation from these matches, we have a selection criterion that matches the overlap ratio of the two transformed foreground infra-red and visible images and select the best ratio, which helps in improving the precision and accuracy and thus describes the best transformation of a video scene.

In future work, we would manage the reservoir attribution for each blob by incorporating the information from a tracker. This would result in even more precise results, in cases where the targets are at different depth planes.

\section*{Acknowledgment}

This work was supported by the FRQ-NT Team research project grant No. 167442.



%



\newpage
{\small
\bibliographystyle{IEEEtran}
\bibliography{IEEEabrv,./bare_conf}
}

\end{document}